\documentclass{article}

\pdfoutput=1

    \PassOptionsToPackage{numbers, compress}{natbib}

\usepackage[preprint]{neurips_2022}

\usepackage[utf8]{inputenc} 
\usepackage[T1]{fontenc}    
\usepackage{hyperref}       
\usepackage{url}            
\usepackage{booktabs}       
\usepackage{amsfonts}       
\usepackage{nicefrac}       
\usepackage{microtype}      
\usepackage{xcolor}         
\usepackage{amsmath}
\usepackage{amssymb}
\usepackage{graphicx}
\usepackage{multicol}
\usepackage{multirow}
\usepackage{color}
\usepackage{transparent}
\usepackage{graphicx}
\usepackage{import}
\usepackage{makecell}

\title{MARS: A Motif-based Autoregressive Model for Retrosynthesis Prediction}

\author{
Jiahan Liu$^{1,2,}$\thanks{Co-first authors. This work is done when Jiahan Liu works as an intern in Tencent AI Lab.}, Chaochao Yan$^{3,}$\footnote[1]{}, Yang Yu$^1$, Chan Lu$^1$, Junzhou Huang$^3$, Le Ou-Yang$^2$, Peilin Zhao$^{1,}$\thanks{Corresponding author.}\\
$^1$ Tencent AI Lab, $^2$ Shenzhen University, $^3$ University of Texas at Arlington\\
\texttt{liujiahan2020@email.szu.edu.cn}, \texttt{chaochao.yan@mavs.uta.edu}, \texttt{kevinyyu@tencent.com} \\
\texttt{chanlu009@gmail.com}, \texttt{jzhuang@uta.edu}, \texttt{leouyang@szu.edu.cn}, \texttt{masonzhao@tencent.com}
}

\begin{document}

\maketitle

\begin{abstract}
  Retrosynthesis is a major task for drug discovery. It is formulated as a graph-generating problem by many existing approaches. Specifically, these methods firstly identify the reaction center, and break target molecule accordingly to generate synthons. 
  Reactants are generated by either adding atoms sequentially to synthon graphs or directly adding proper leaving groups. 
  However, both two strategies suffer since adding atoms results in a long prediction sequence which increases generation difficulty, while adding leaving groups can only consider the ones in the training set which results in poor generalization.
  In this paper, we propose a novel end-to-end graph generation model for retrosynthesis prediction, which sequentially identifies the reaction center, generates the synthons, and adds motifs to the synthons to generate reactants.
  Since chemically meaningful motifs are bigger than atoms and smaller than leaving groups, our method enjoys lower prediction complexity than adding atoms and better generalization than adding leaving groups.
  Experiments on a benchmark dataset show that the proposed model significantly outperforms previous state-of-the-art algorithms.
\end{abstract}

\section{Introduction}
Retrosynthesis prediction plays an important role in synthesis planning and drug discovery. Its purpose is to predict physically available reactants that can synthesize target molecules. Corey \cite{corey1991logic} first proposed the concept of retrosynthesis, which became a hot research topic in the field of organic chemistry. However, there are around \(10^7\) reactions and molecules in the published synthetic-organic knowledge \cite{gothard2012rewiring}. Due to the huge chemical search space, the large number of possible combinations make this task very challenging. In the past, potential reactants were derived by experienced chemists, but it was inefficient and heavily reliant on the chemists' experience and knowledge. For example, the complete synthetic route of vitamin b12 was explored by Woodward \cite{woodward1973total} in conjunction with hundreds of chemists, and it took 11 years. To improve efficiency, chemists have turned to computer-aided synthesis planning (CASP) tools to design synthetic pathways. Several rule-based systems \cite{kayala2012reactionpredictor,marcou2015expert} have been developed and achieve excellent results for specific reaction types, but they suffer from high complexities and poor generalization.

With the development of deep learning, deep models have spawned a series of promising proposals, greatly increasing the efficiency of synthetic route design. These models are widely classified into two types: template-based \cite{segler:neuralsym, coley:retrosim, dai:gln, yan2021retrocomposer} and template-free \cite{liu:seq2seq, zheng:scrop, shi:g2gs, ydz:retroxpert, sun2021towards}. Template-based models rely on templates that are manually extracted by experienced chemists, or automatically extracted from large-scale data. The core task of these methods is to match the product and the reactants to the appropriate template, which reflects the reaction center of the target molecule in a certain type of reaction. Template-based methods are highly interpretable and overcome the problem that previous rule-based systems give conflicting results with functional groups \cite{segler:neuralsym}. However, they are limited by expensive subgraph matching and difficult scaling.

\begin{figure}[t]
\centering
\includegraphics[width=\textwidth]{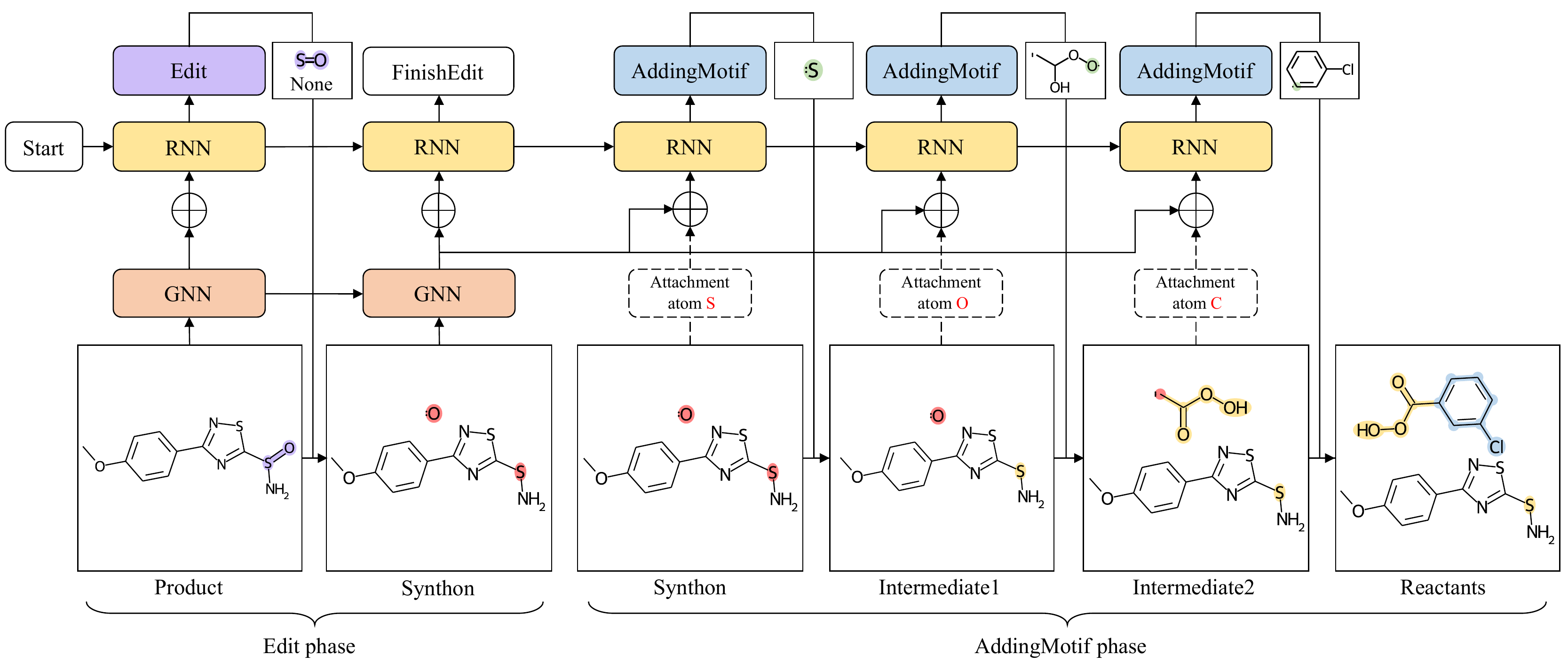}
\caption{Reactants generation procedure of the proposed MARS.
\emph{Edit} and \emph{AddingMotif} indicate graph transformation actions, where the \emph{Edit} phase describes bond and atom changes from product to synthons and plays the role of reaction center identification while \emph{AddingMotif} phase conducts synthon completion by adding proper motifs to synthons. Input molecular graphs are encoded by the Graph Neural Network (GNN), and the Recurrent Neural Network (RNN) predicts graph transformation operations sequentially. In \emph{Edit} phase, the RNN predicts a sequence of \emph{Edit} operations until the \emph{FinishEdit} which indicates the end of \emph{Edit} phase as well as the start of  \emph{AddingMotif} phase. In \emph{AddingMotif} phase, the RNN adds motifs sequentially until no attachment atoms (highlighted in pink) remained. In the above example, the first \emph{Edit} operation applies to the S=O bond, and the new bond type is None which indicates removing the bond. For the \emph{AddingMotif} operation, the \emph{interface-atom} (green) in the motif and the \emph{attachment atom} in the synthon/intermediate represents the same atom and are merged into single atom when attaching the motif to the synthon/intermediate. }
\label{model}
\end{figure}

Template-free methods can be generally divided into sequence-based and graph-based methods. Sequence-based methods treat retrosynthesis prediction as a machine translation task. These methods use an encoder-decoder model (e.g. LSTM and Transformer) \cite{liu:seq2seq, kgt:transformer, zheng:scrop} to translate SMILES\footnote{https://www.daylight.com/dayhtml/doc/theory/theory.smiles.html} sequences of target molecules into reactants SMILES sequences  without atom mapping and subgraph matching.
Although sequence-based methods can implicitly learn reactive rules and easily scale to larger datasets, they ignore the rich topological information in molecular graphs and easily lead to invalid reactant molecules. 
Recently, many graph-based works for retrosynthesis have been proposed with the development of graph neural networks. 
These methods usually follow the similar paradigm: i) reaction center identification and ii) synthon completion. G2Gs \cite{shi:g2gs}, RetroXpert \cite{ydz:retroxpert}, and GraphRetro \cite{srb:graphretro} all use a two-stage framework to formulate above two subtasks. However, due to the different optimization objectives of the two separate models, two-stage methods are difficult to achieve optimal results and have poor generalization. Additionally, GraphRetro uses leaving groups to complete synthons. However, the unreasonable design of the leaving groups make the samples unbalanced and the model suffers from low generalization. MEGAN \cite{sbb:megan} constructs an end-to-end model but completes synthons with tiny units like single atoms and benzene, while the lengthy prediction process makes the reactant generation challenging.

In this work, we propose a novel \textbf{M}otif-based \textbf{A}utoregressive model for \textbf{R}etro\textbf{S}ynthesis prediction (\textbf{MARS}). 
The workflow of the entire model is shown in \autoref{model}. To explore the potential relationship between reaction center identification and synthon completion, we design a end-to-end graph generation framework to jointly tackle these two subtasks. Instead of using single atom or ring to completing synthons, we employ a predefined motif vocabulary from training reactions. 
Motifs are fine-grained components that enjoys lower redundancy, more balanced data distribution, and more generative flexibility than leaving groups proposed by GraphRetro \cite{srb:graphretro}. 
We describe each step from product to reactants through carefully designed graph editing actions represented as a complete transformation path, then adapt the RNN to learn to generate a transformation path in an autoregressive manner. Our main contributions can be summarized as:
\begin{itemize}
    \item We integrate the two subtasks of reaction center identification and synthon completion into a unified framework, and adapt an encoder-decoder architecture for retrosynthesis prediction to train the model in an end-to-end manner.
    \item We extract a chemically meaningful motif vocabulary from training reactions without additional chemical knowledge, which greatly improves generalization ability.
    \item Experiments on the benchmark dataset show that our model achieve the state-of-the-art retrosynthesis performance with a top-\(1\) accuracy of \(54.6\%\) and \(66.2\%\) when w/o and w/ reaction class, respectively. 
\end{itemize}

\section{Related works}

\paragraph{Molecular Graph Generation Methods} Various molecular graph generation methods have been proposed with the aim of generating chemically valid molecules with specific chemical properties. MolGAN \cite{de2018molgan} generates molecules via generative adversarial networks. JT-VAE \cite{jbj:junctiontree} first breaks a molecular graph into several disconnected subgraphs and then designs a junction tree variational autoencoder for molecule generation. Recently, much attention has been drawn to autoregressive based models. GCPN \cite{you2018gcpn} formulates molecular graph generation as a Markov Decision Process. MolecularRNN \cite{popova2019molecularrnn} utilizes a recurrent neural network to generate the nodes and edges. GraphAF \cite{shi2020graphaf} designs a flow-based autoregressive model to dynamically generate nodes and edges based on history subgraph structures. Our proposed method can be considered as a conditional molecular graph generation method based on an autoregressive model.

\paragraph{Template-free Retrosynthesis Prediction Methods}
Template-free methods are data-driven methods that can be divided into sequence-based and graph-based methods. Inspired by natural language processing (NLP), the researchers regard the conversion of products to reactants as a machine translation problem, since molecules can be represented as SMILES strings. For example, Seq2seq \cite{liu:seq2seq} and Transformer \cite{kgt:transformer} simply apply machine translation models to retrosynthesis tasks, resulting in the generation of ineffective molecules. To remedy the grammatically incorrect output in the previous models, SCROP \cite{zheng:scrop} combines the standard Transformer with a grammar corrector. Although these methods simplify retrosynthetic models, they ignore the rich structural information in molecular graph and are poorly interpretable.

Graph-based approaches model the retrosynthesis task as two steps: i) break the target molecule into incomplete molecules called synthons, and then ii) complete them into reactants using subgraph units such as atoms or leaving groups. G2Gs \cite{shi:g2gs}, RetroXpert \cite{ydz:retroxpert}, and GraphRetro \cite{srb:graphretro} build two independent models to implement the above steps respectively. MEGAN \cite{sbb:megan} constructs an end-to-end graph generative model while completes synthon with individual atoms and benzene. Our work is closely related to graph-based models, but fundamentally different from above methods. First, rather than treating reaction center 
identification and synthon completion as two completely independent subtasks like \cite{shi:g2gs, ydz:retroxpert, srb:graphretro}, our work integrates these two subtasks into an end-to-end framework. Second, compared to the high prediction complexity of completing synthon with small units \cite{sbb:megan}, adding motifs to synthons can greatly reduce the length of the prediction sequence. It is worth noting that motifs are distinct from leaving group proposed by \cite{srb:graphretro} and the differences are discussed in Section 3.1.

\section{Proposed method}

In this section, we first describe the construction of the transformation path, and then detail our proposed model MARS.

\paragraph{Notations} Molecules are represented as graph \(G=(V,E)\) with \(n\) atoms and \(m\) bonds, where \(V\) is the set of atoms (nodes) and \(E\) is the set of bonds (edges). 
Each atom \(u\) has a feature vector \(\boldsymbol{x}_u\) indicating its atom type, degree, chiral tag, the number of hydrogen and so on. Similarly, each bond \((u,v)\) has a feature vector \(\boldsymbol{x}_{u,v}\) indicating bond type, stereo, aromaticity and so on (See Appendix \ref{atom_bond_features} for details). 
All features are computed by the rdkit\footnote{https://www.rdkit.org} package. For convenience, an index \(i\) is assigned to each bond and atom, where the bond index is its index in rdkit, and the  atom index is its index in rdkit plus \(m\). 
In addition, each bond has a 4-dimensional one-hot vector \(\boldsymbol{r}_b\) representing its bond type, including none, single, double and triple bonds, respectively. All bonds and atoms have a label \(s_i\in \{0, 1\}\) indicating whether they belong to the reaction center.

\begin{figure*}[t]
\begin{center}
\centerline{\includegraphics[width=\textwidth]{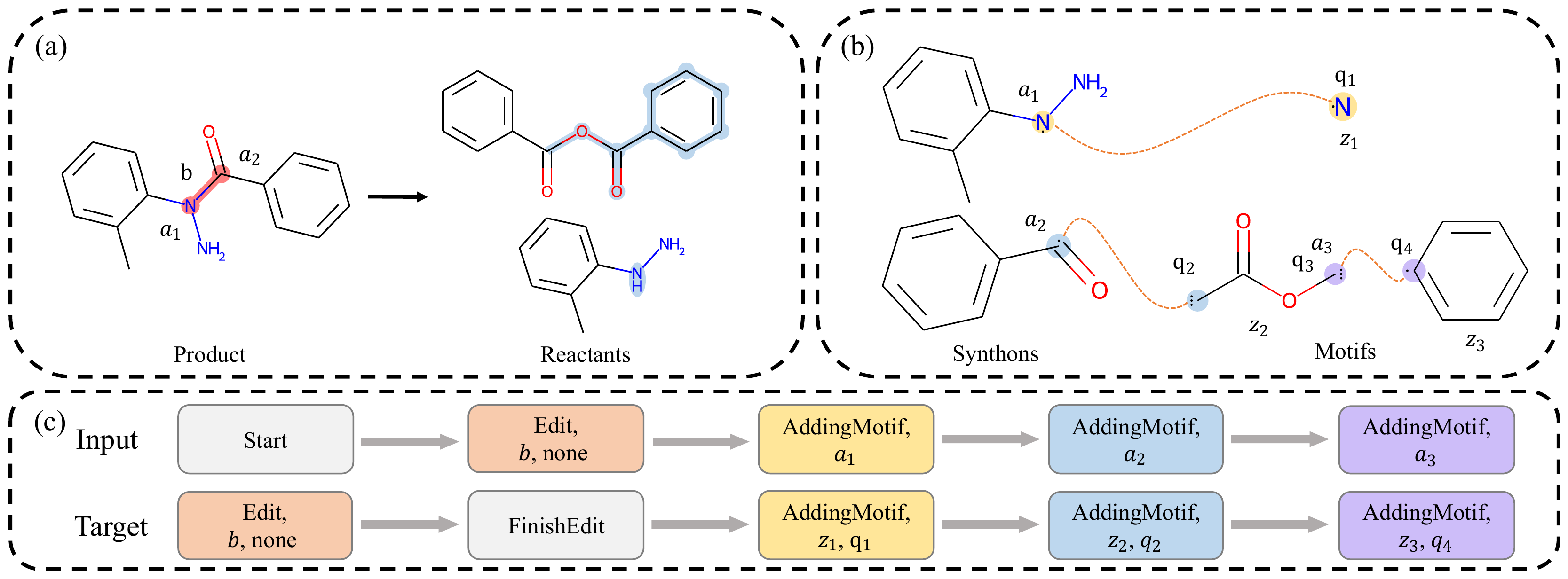}}
\caption{(a) The conversion of products to reactants. The bond \(b\) marked pink represents the reaction center, while the atoms \(a_1\) and \(a_2\) are attachment atoms; (b) The junction tree constructed by synthons and motifs \(z\). The atoms of the same color represent connected attachment atom and interface atom \(q\), which are the same atom in reactant. The arrows represent the parent node pointing to the child node. After connecting synthons with \(z_2\), the interface atom \(q_3\) becomes the attachment atom \(a_3\); (c) The input and target transformation paths for training RNN, constructed according to (a) and (b).}
\label{junctiontree}
\end{center}
\vskip -0.2in
\end{figure*}

\subsection{Transformation Path Construction}

In MARS, we formulate the retrosynthesis prediction as a graph generation problem. Specifically, a sequence is predicted to transform a given product to reactants, in which each token represents a graph editing action. 
Therefore, we pre-construct a transformation path for each product to describe the conversion from product to reactants. 
A transformation path contains two parts: \emph{Edit} phase and \emph{AddingMotif} phase (\autoref{model}). \emph{Edit} phase plays the role of reaction center identification and it describes bond and atom changes from product to synthons, while \emph{AddingMotif} phase conducts synthon completion by adding proper pre-defined motifs to synthons. 
In particular, we introduce a hierarchical structure to represent the connection between synthons and motifs \autoref{junctiontree}b, named junction tree \cite{jbj:junctiontree}, which provides an efficient way to create \emph{AddingMotif} sequences.
In order to integrate \emph{Edit} and \emph{AddingMotif} actions into a complete transformation path, four graph transformation tokens are defined to represent the graph transformation: \emph{Start}, \emph{Edit}, \emph{FinishEdit}, and \emph{AddingMotif}. 
Except for auxiliary actions \emph{Start} and \emph{FinishEdit}, each token in transformation path contains three parts: edit action \(\pi\), edit object \(o\), and edit state \(\tau\). 
We then elaborate the representation of the reaction center in the Edit sequence, motif extraction, and junction tree construction.

\paragraph{Edit Sequence Construction} \emph{Edit} includes {bond edit} and {atom edit} and it describes bond and atom edits applied to a target to obtain synthons. A bond edit is to add or remove a bond, or to change the bond type between two heavy atoms, while an atom edit is to change the number of hydrogen or charge of an atom.
We refer to the atoms at both ends of the changed bond, as well as the atoms with changed hydrogen or charges as \emph{attachment atoms}, since we need attach a motif to every attachment atom to complete synthons.
In this way, the reaction center of the product can be encoded as an \emph{Edit} sequence. Each \emph{Edit} token in the sequence is a tuple of \emph{(edit action, edit object, edit state)}.
For example, in \autoref{junctiontree}a, the bond marked pink is the reaction center of the given product. The \emph{Edit} tuple is denoted as \emph{(Edit, b, none)}, where \emph{b} is the bond index and \emph{none} describes the new bond type which indicates removing the bond.

Once we obtain synthons resulting from the \emph{Edit} operations, we can add motifs to synthons to generate reactants. \emph{AddingMotif} token represents adding an motif from the vocabulary and attach the motif to a specific attachment. 
A motif also contains one or more attachments. We adds motifs sequentially until all attachments are properly processed. 

\paragraph{Motif Extraction}
Through  \emph{Edit}, a product molecular graph is decomposed into a group of incomplete subgraphs, named synthons. Combining appropriate motifs and attachments, synthons can be reconstructed to valid reactant molecular graphs. In other words, motifs are the subgraphs of reactant molecular graph. The details of motif extraction are summarised as follows.
\begin{itemize}
\item Break the bonds that connect the synthon in the reactants to get a set of subgraphs. Each subgraph retains the attachment atom connected to it on the synthon, resulting in a coarse-grained motif.
\item If two connected atoms belong to two rings, the bond between them will be broken, resulting in two independent motifs.
\item If one atom belongs to a ring and the other has a degree greater than 1, the bond between them will be broken, resulting in two independent motifs.
\end{itemize}
Finally we obtain a motif vocabulary \(Z\) of size  \(|Z|=210\) from the USPTO-50K training set \cite{schneider2016s}.
It is worth noting that motifs are fundamentally different from leaving groups proposed by previous method \cite{srb:graphretro}. i) A motif is connected to an attachment atom, while a leaving group is combined with a synthon. Since one synthon may contains more than one attachment, a leaving group may consist of multiple disconnected subgraphs (i.e. motifs). ii) Motif retains the corresponding attachment atom on synthon, called the \emph{interface-atom}. We notice that a large part of the added leaving groups is single hydrogen, which makes the frequency of leaving groups extremely unbalanced. iii) A large leaving group may contain multiple rings or large branched chains, appearing infrequently in the dataset. We cut it into multiple small motifs that are common in the dataset to reduce redundancy. 

\paragraph{Junction Tree Construction}
Motivated by the chemical intuition, reactants are considered that can be decomposed into synthons and motifs, where synthons are the molecule fragments formed by breaking the bond in the product and motif is a subgraph of reactants. Here, junction tree is introduced to maintain the connection between synthons and motifs according to \cite{jbj:junctiontree}. 
A junction tree represents synthons and motifs as a hierarchical tree structure, where the group of synthons is set as the root node and motifs are set as children nodes (\autoref{junctiontree}b). The connected edge between two nodes indicates that they are directly connected in the reactant, denoted as \emph{(attachment, motif, interface-atom)}. The trees are traversed using depth-first search (DFS) to preserve the linked edges between nodes, obtaining the training input and target AddingMotif paths. In input path, each token contains an action \emph{AddingMotif} and an object \emph{attachment index}, while for target path, the object consists of the \emph{motif index} \(z\) and the \emph{interface-atom index} \(q\). 

By combining the above Edit sequence with the AddingMotif path, as well as other auxiliary actions, we obtain the input and target transformation paths corresponding to each product (\autoref{junctiontree}c).

\subsection{Graph Encoder}
Graph neural networks (GNNs) \cite{kipf2016semi, hamilton2017inductive, velivckovic2017graph, shi:graphtransformer} are a series of neural network architectures acting on graph structure and properties, which update the representation vectors (i.e. embeddings) of nodes through a message passing mechanism. We encode the latent representation of \(G\) by an \(L\)-layer message passing neural network, denoted as \(\operatorname{MPNN}(\cdot)\). The atom representations \(\{\boldsymbol{h}_v\in \mathbb{R}^{D}|v\in G\}\) and bond representations \(\{\boldsymbol{h}_{v,u}\in \mathbb{R}^{D}|{u\in{\mathcal{N}(v)},v\in G}\}\) can be generally expressed as follows
\begin{align}
    &\boldsymbol{h}_v^{L} = \operatorname{MPNN}(G, \boldsymbol{x}_v, \{\boldsymbol{x}_{v,u}\}_{u\in{\mathcal{N}(v)}}),\\
    &\boldsymbol{h}_{v,u} = \operatorname{MLP}_{bond}(\boldsymbol{h}_v^{L} \ \|\ \boldsymbol{h}_u^{L}), u\in{\mathcal{N}(v)},
\end{align}
where \(\mathcal{N}(v)\) denotes the neighbors of atom \(v\), \(\|\) is the concatenation operation and \(\operatorname{MLP}(\cdot)\) is Multi-Layer Perceptrons. Additionally, to facilitate subsequent tasks, we represent atoms in the same form as the bond representation:
\begin{align}
    \boldsymbol{h}_{v,v} &= \operatorname{MLP}_{bond}(\boldsymbol{h}_v^{L} \ \|\ \boldsymbol{h}_v^{L}).
\end{align}
For simplicity, we denote bonds and atoms of the same form of representation as \(\boldsymbol{e}_i\in \{h_{v,u}\}_{u\in \mathcal{N}(v)\bigcup v}\) where \(i\) is the index of bonds and atoms. 
The final graph representations \(\boldsymbol{h}_G\) is defined by aggregating the whole atom representations using a global attention pooling function \cite{li2015gated}. Similarly, the graph representation of synthons \(\boldsymbol{h}_{syn}\) can also be computed.

\subsection{Autoregressive Model}
Motivated by \cite{popova2019molecularrnn, shi2020graphaf}, we formulate retrosynthesis prediction as an autoregressive-based conditional molecule generation problem. Given a product graph \(G_P\), the autoregressive model generates a new graph structure \(G_t\) based on the incomplete graph of the previous steps, until the reactant graph \(G_R\) is finally obtained. The general process can be defined as a jointly conditional likelihood function:
\begin{align}
    P(G_R) = \prod_{t=1}^{N}P(G_t|G_0,\ldots,G_{t-1})
    =\prod_{t=1}^{N}P(G_t|G_{<t}),
    \label{conditional_likelihood_function1}
\end{align}
where \(N\) is the length of generated sequence and \(G_0\) is \(G_P\). To be clear, the intermediate graph structure \(G_t\) is not directly generated by the model. Instead, the graph editing action \(\pi\), edit object \(o\) (i.e. bond, atom or motif) and its edit state \(\tau\) (e.g. new bond type or interface-atom) are generated from the historical graph editing sequence, and then applied to \(G_{t-1}\) to obtain a new graph structure. That is, given the historical edited objects, edited states and incomplete graphs, the likelihood in \autoref{conditional_likelihood_function1} can be modified as:
\begin{align}
    \label{conditional_likelihood_function2}
    P(G_R) = \prod_{t=1}^{N}P(\pi_{t},o_{t},\tau_{t}|o_{<t},\tau_{<t},G_{<t}).
\end{align}
A recurrent neural network (RNN) is employed to model \autoref{conditional_likelihood_function2} by encoding object, state and incomplete graph of last step to decode the output \(\boldsymbol{u}_{t}\in \mathbb{R}^{D}\). To incorporate the global topological information of \(G_P\) in generation process, we concatenate \(\boldsymbol{h}_{G}\) and \(\boldsymbol{u}_{t}\) for subsequent prediction. 
\begin{gather}
    \boldsymbol{u}_{t}=\operatorname{GRU}(\operatorname{input}_{t-1}, \operatorname{hidden}_{t-1}),\ where\  \operatorname{input}_{0}=\boldsymbol{0},\ 
    \operatorname{hidden}_{0}=\sigma_{G}(\boldsymbol{h}_{G}),\\
    \boldsymbol{\psi}_{t} = \boldsymbol{h}_{G} \ \|\  \boldsymbol{u}_{t},
    \label{gru-eq}
\end{gather}
where \(\operatorname{GRU}(\cdot)\) is Gated Recurrent Unit (GRU) \cite{chung:gru,li:gru}, \(\operatorname{input}_{t}\in \mathbb{R}^{D}\) and \(\operatorname{hidden}_{t}\) are the input embedding and hidden state of GRU at step \(t\), and \(\sigma(\cdot)\) is the embedding function. The generation process begins at action \emph{Start}, and at each step \(t\), we generate graph editing action \(\hat{\pi}_{t}\) by follow:
\begin{equation}
    \begin{aligned}
    &\hat{\pi}_{t}=\operatorname{softmax}(\operatorname{MLP}_{act}(\boldsymbol{\psi}_{t})).
    \label{actions}
    \end{aligned}
\end{equation}
\paragraph{Edit Phase} When the predicted action is \emph{Edit}, the process reaches Edit phase. At step \(t\), the model firstly assigns an editing score \(\hat{s}_i\) to each bond and atom, indicating the likelihood that the bond or atom is considered an edit object. The selected atom or atoms at both ends of the selected bond are set as attachments. The model predicts the new bond type \(\hat{r}_b\) for the edit object. Then we apply the edit object and its new bond type to modify synthon structure and perform \(\operatorname{MPNN}(\cdot)\) to obtain synthon embedding \(\boldsymbol{h}_{t}^{syn}\). Finally, \(\operatorname{input}_{t}\) is updated by synthon embedding, edit object and its new bond type.
\begin{equation}
    \begin{aligned}
    &\hat{s}_{i}=\operatorname{sigmoid}(\operatorname{MLP}_{target}(\boldsymbol{\psi}_{t} \ \|\ \sigma_e(\boldsymbol{e}_{i}))),\\
    &\hat{r}_{b}=\operatorname{softmax}(\operatorname{MLP}_{type}(\boldsymbol{\psi}_{t} \ \|\ \sigma_e(\boldsymbol{e}_{\mathop{\arg\max}\limits_{i}(\hat{\boldsymbol{s}}_{i})}))),\\
    &\operatorname{input}_{t}=\boldsymbol{h}_{t}^{syn} + \sigma_e(\boldsymbol{e}_{\mathop{\arg\max}\limits_{i}(\hat{\boldsymbol{s}}_{i})}) + \sigma_b(\hat{r}_{b}).
    \end{aligned}
\end{equation}
When the predicted action is \emph{FinishEdit}, Edit phase ends and AddingMotif phase begins. The synthon structure is fixed and its embedding is denoted as \(\boldsymbol{h}_{syn}\). The attachments \(\{a\}\) are sorted according to their atom indexes in \(G_P\), and \(\operatorname{input}_{t}\) is updated:
\begin{align}
    \operatorname{input}_{t} = \boldsymbol{h}_{syn} + \sigma_{atom}(a_t),\ where\ a_t \in \{a\}.
    \label{update_input_motif_adding}
\end{align}
\paragraph{AddingMotif Phase} In this phase, the model traverses all attachments \(\{a\}\) sequentially and assigns a suitable motif to each attachment. Motif prediction is treated as a multi-classification task on the motif vocabulary \(Z\). Once obtaining motif \(\hat{z}\), the model needs to determine which interface-atom \(\hat{q}\) on the motif corresponds to the attachment atom \(a_t\). Thus, motif \(\hat{z}\) and interface-atom \(\hat{q}\) are predicted as follows:
\begin{equation}
    \begin{aligned}
    &\hat{z}=\operatorname{softmax}(\operatorname{MLP}_{motif}(\boldsymbol{\psi}_{t})),\\
    &\hat{q}=\operatorname{softmax}(\operatorname{MLP}_{interface}(\boldsymbol{\psi}_{t} \ \|\ \sigma_z(\hat{z}))).
    \end{aligned}
\end{equation}
If the predicted motif \(\hat{z}\) only contains one interface-atom, \(\operatorname{input}_{t}\) is computed as \autoref{update_input_motif_adding}, while if \(\hat{z}\) contains multiple interface-atoms, \(\operatorname{input}_{t}\) is updated as follows:
\begin{equation}
    \begin{aligned}
    \operatorname{input}_{t}=\boldsymbol{h}_{syn}+\sigma_z(\hat{z})+\sigma_{atom}(a_t).
    \end{aligned}
\end{equation}
Please kindly note that it is not necessary to incorporate an action to indicate the end of the process. The generation process ends until all attachments on the synthons and added motifs have been traversed. Finally, a transformation path is generated and applied on the product to obtain reactants.

\subsection{Training and Inference}
\paragraph{Training}
The objective of our model is to predict target transformation path given training transformation path. We optimize the prediction of new types, motifs and interface-atom indexes by cross-entropy loss \(\mathcal{L}_c\), and the prediction of reaction centers by binary cross-entropy loss \(\mathcal{L}_b\). Thus, the overall loss function for MARS is:
\begin{equation}
    \begin{aligned}
    \mathcal{L}\!=\!&\sum^{N_1+N_2}{\mathcal{L}_c(\hat{a}, a)}\!+\!
    \sum^{N_1}{\Bigg[\sum_{i=0}^{n+m-1}\mathcal{L}_b(\hat{s}_{i},s_{i})\!+\!
    \mathcal{L}_c(\hat{r}_{b}, r_{b})\Bigg]}\!+\!
    \sum^{N_2}{\Bigg[\mathcal{L}_c(\hat{z},{z})\!+\!
    \mathcal{L}_c(\hat{q}, q)\Bigg]},
\end{aligned}
\end{equation}
where \(N_1\) and \(N_2\) denote the lengths of Edit and AddingMotif sequences respectively. To reduce convergence difficulties, we adopt an efficient strategy teacher-forcing \cite{williams1989learning} to train our model. When training model, the strategy utilizes ground truth as the input for the model instead of directly uses the output from the last time step as the input at current time step. 

\paragraph{Inference}
We use beam search \cite{tillmann2003word} with hyperparameter \(k\) to rank the predictions. At each time step, the top-\(k\) best results are selected as the input at next time step based on the log-likelihood score function. In other words, the process can be described as the construction of a search tree, in which the leaf nodes with highest scores are expanded with their children nodes while the other leaf nodes are dropped. Please kindly note that atom-mapping in testing set is unnecessary in inference phase. 

\section{Results}

\paragraph{Data} We primarily test the efficiency of our approach using a benchmark dataset called USPTO-50K \cite{schneider2016s}, which comprises 50K reactions from the US patent literature of \(10\) various classes (See Appendix \ref{data_information} for details). We use the dataset and the same training/ validation/ testing splits in \(8\):\(1\):\(1\) as \cite{coley:retrosim, dai:gln}. It has been reported that the USPTO dataset contains a shortcut in \(75\%\) of the product molecules, i.e. the atom of atom-mapping "1" is part of the reaction center. These shortcuts are eliminated by canonicalizing product SMILES and reassigning atom-mapping to reactant atoms.

\paragraph{Evaluation} The evaluation metric we use is top-\(k\) accuracy, defined as the percentage of the ground truth reactants in the top-\(k\) suggestions of our model. The accuracy is calculated by comparing the predicted reactants with ground truth in canonical SMILES format.

\paragraph{Experimental setting} We use PyTorch \cite{paszke2019pytorch} and PyG \cite{fey2019pyg} to implement our model. We use Graph Transformer \cite{shi:graphtransformer} as our graph encoder, which stacked six eight-head attentive layers. The GRU network is implemented with three layers. Embedding size \(D\) in our model is set to \(512\).

In all experiments, we train on USPTO-50K for 100 epochs, using a batch size of \(32\) and the Adam \cite{kingma2014adam} optimizer with initial learning rate of \(0.0003\). 
The training on USPTO-50K takes approximately \(17\)h on a single NVIDIA Tesla V100 GPU. The beam size \(k\) is set to \(10\) in inference phase. (See Appendix \ref{experimental_setting} for details)

\paragraph{Baseline} We take three template-based and seven template-free methods as our competitors. All results are derived from their original reports, except for NeuralSym \cite{segler:neuralsym} reported by GLN \cite{dai:gln}, and corrected results reported by RetroXpert\cite{ydz:retroxpert} on their website \footnote{https://github.com/uta-smile/RetroXpert}. The details are reported in Appendix \ref{baselines}.

\subsection{Overall Performance}
We present the top-\(k\) accuracy in \autoref{TopK-acc}, where N ranges from \(\{1,3,5,10\}\). We evaluate both the reaction class unknown and reaction class known. 

\begin{table}[ht]
\caption{Top-\(k\) accuracy for retrosynthesis prediction on USPTO-50K.}
\label{TopK-acc}
\begin{tabular}{@{}llcccccccc@{}}
\toprule
\multicolumn{2}{c}{\multirow{2}{*}{Methods}}     & \multicolumn{8}{c}{Top-n Accuracy (\%)}                                                                                       \\ \cmidrule(l){3-10} 
\multicolumn{2}{c}{}                             & \multicolumn{4}{c}{Reaction Type Known}                       & \multicolumn{4}{c}{Reaction Type Unknown}                     \\ \midrule
\multicolumn{2}{l}{}                             & 1             & 3             & 5             & 10            & 1             & 3             & 5             & 10            \\ \midrule
\multirow{3}{*}{Template-based} & RetroSim       & 52.9          & 73.8          & 81.2          & 88.1          & 37.3          & 54.7          & 63.3          & 74.1          \\
                                & NeuralSym      & 55.3          & 76            & 81.4          & 85.1          & 44.4          & 65.3          & 72.4          & 78.9          \\
                                & GLN            & 64.2          & 79.1          & 85.2          & 90            & 52.5          & 69            & 75.6          & 83.7          \\ \midrule
\multirow{3}{*}{Sequence-based} & SCROP          & 59            & 74.8          & 78.1          & 81.1          & 43.7          & 60            & 65.2          & 68.7          \\
                                & LV-Transformer & \textbf{-}    & \textbf{-}    & \textbf{-}    & -             & 40.5          & 65.1          & 72.8          & 79.4          \\
                                & DualTF         & 65.7          & 81.9          & 84.7          & 85.9          & 53.6          & 70.7          & 74.6          & 77            \\ \midrule
\multirow{5}{*}{Graph-based}    & G2Gs           & 61            & 81.3          & 86            & 88.7          & 48.9          & 67.6          & 72.5          & 75.5          \\
                                & RetroXpert     & 62.1          & 75.8          & 78.5          & 80.9          & 50.4          & 61.1          & 62.3          & 63.4          \\
                                & MEGAN          & 60.7          & 82            & 87.5          & 91.6          & 48.1          & 70.7          & 78.4          & 86.1          \\
                                & GraphRetro     & 63.9          & 81.5          & 85.2          & 88.1          & 53.7          & 68.3          & 72.2          & 75.5          \\
                                & MARS (Ours)           & \textbf{66.2} & \textbf{85.8} & \textbf{90.2} & \textbf{92.9} & \textbf{54.6} & \textbf{76.4} & \textbf{83.3} & \textbf{88.5} \\ \bottomrule
\end{tabular}
\end{table}

\paragraph{Reaction class unknown}
When reaction class is unknown, our model surpasses both template-based and template-free models. Our model outperforms GraphRetro by \(0.9\%\) and MEGAN by \(6.5\%\) in terms of top-\(1\) accuracy. For larger \(k\), our model still enjoys high performance, which is over \(8.1\%\) than GraphRetro and \(2.4\%\) than MEGAN. We notice that both our model and MEGAN outperform two-stage models when \(n\geq 3\). Benefiting from an end-to-end model, our model is able to explore the underlying relationship between reaction centers and synthon completion, rather than relying on two separate modules with different optimization objectives. In addition, our model takes advantage of motifs that are larger than individual atoms and smaller than leaving groups. This allows our model to avoid the high complexity of long prediction sequences without losing flexibility.

\paragraph{Reaction class known}
When reaction class is given, our model outperforms MEGAN and GraphRetro by \(5.5\%\) and \(2.3\%\) in top-1 accuracy. For larger \(k\), our model also achieves state-of-the-art top-\(k\) accuracy of \(85.8\%\), \(90.2\%\) and \(92.9\%\), which is over \(4.3\%\) higher than GraphRetro. Even though template-based methods can exploit the reaction class to narrow the template space and improve the accuracy, they also suffer from poor generalization. On the contrary, our model can improve accuracy while still maintaining high generalization performance. 

\subsection{Ablation Study of Synthon Embedding}

To understand the importance of synthon embedding, we conduct ablation study by removing it in autoregressive model. As shown in \autoref{tab:ablation_without_synthon}, when the synthon embedding is not included, the top-\(1\) accuracy drops by \(4.9\%\) if given the reaction class and \(10.5\%\) if not given. This demonstrate that synthon embedding plays indispensable roles in the generation process. We observe that synthon structure information helps model determine \emph{FinishEdit} action, while the model without synthon embedding suffers from the problem of repeatedly predicting edit objects in Edit phase.

\begin{table}[ht]
\centering
\caption{Top-\(k\) accuracy of synthon embedding ablation study. \emph{MARS-w/o synthon} indicates MARS without synthon embedding.}
\label{tab:ablation_without_synthon}
\begin{tabular}{@{}lcccccccc@{}}
\toprule
\multicolumn{1}{c}{\multirow{3}{*}{Method}} & \multicolumn{8}{c}{Top-k Accuracy (\%)}                                                                                       \\ \cmidrule(l){2-9} 
\multicolumn{1}{c}{}                        & \multicolumn{4}{c}{Reaction Class Known}                      & \multicolumn{4}{c}{Reaction Class Unknown}                  \\ 
\cmidrule(l){2-9} 
                                            & 1             & 3             & 5             & 10            & 1             & 3             & 5             & 10            \\ \midrule
MARS-w/o synthon                         & 61.3          & 73.5          & 76.3          & 81.8          & 44.1          & 58.5          & 63.0          & 69.3          \\ 
MARS                                 & \textbf{66.2} & \textbf{85.6} & \textbf{90.2} & \textbf{92.9} & \textbf{54.6} & \textbf{76.4} & \textbf{83.3} & \textbf{88.5} \\ \bottomrule
\end{tabular}
\end{table}

\subsection{Prediction Visualization}

\begin{figure*}[ht]
\begin{center}
\centerline{\includegraphics[width=0.95\textwidth]{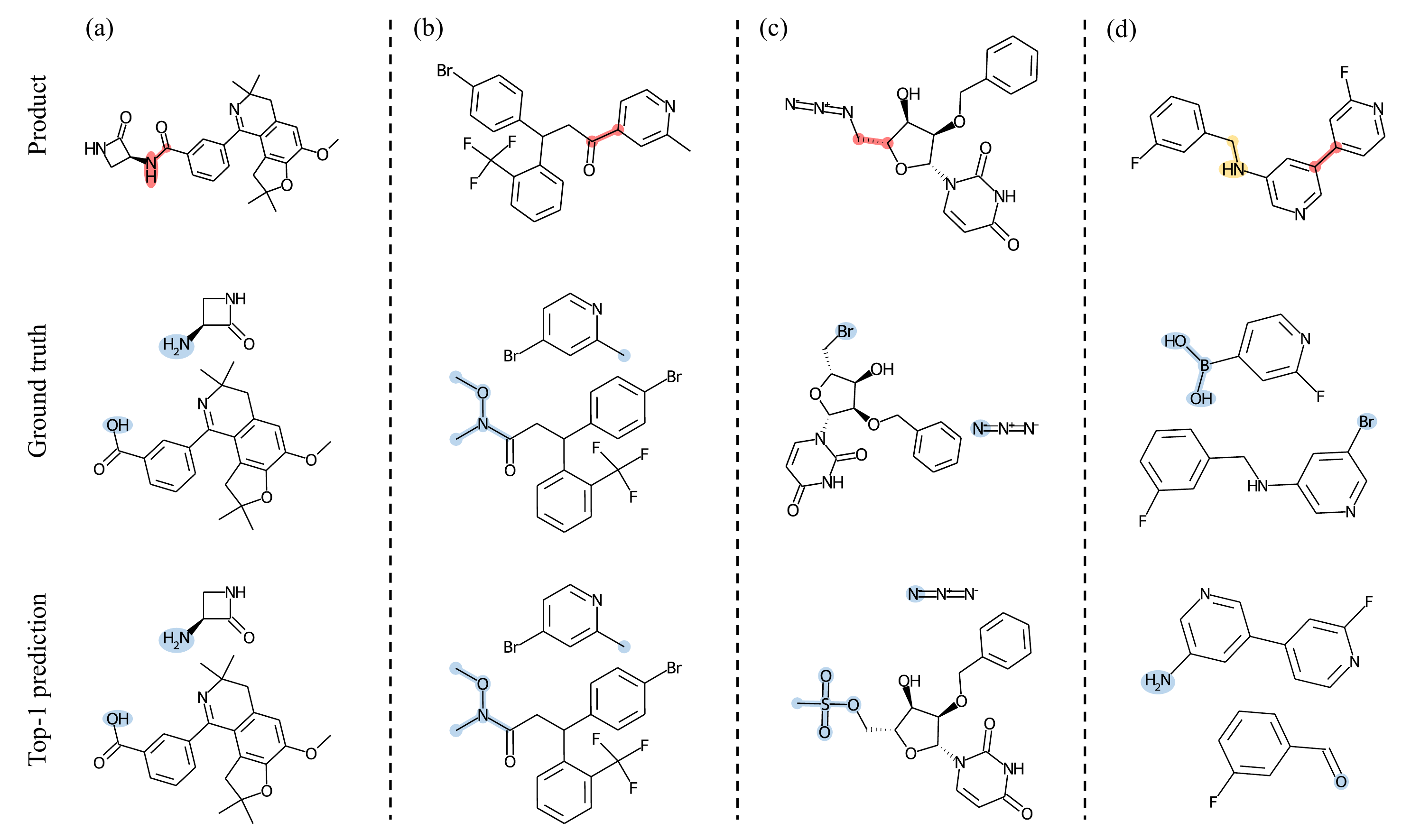}}
\caption{Example Predictions. \emph{Red} indicates the correct reaction centers while \emph{yellow} represents the error one predicted by our model, and \emph{blue} indicates the added motifs. (a)(b). Examples of successful predictions by our model. (c). Correctly predicted reaction center but added wrong motif. (d). Incorrectly predicted reaction center.}
\label{examples}
\end{center}
\vskip -0.25in
\end{figure*}

To better illustrate the prediction performance of our model, we visualize four ground truth reactants and top-1 predicted reactants from USPTO-50K test set in \autoref{examples}. \autoref{examples}a and 3b show the two correct reactants predicted by our model. It can be seen that our model can accurately predict the reaction centers and add the appropriate motifs. Compared to methods that add atom by atom or benzene, our model is insensitive to the size of motifs. This means that our model is able to accurately assign the correct motifs for the synthons. \autoref{examples}c shows a failure in which the reaction center is correctly predicted but motifs are different from the ground truth. However, the predicted reactants are chemically reasonable, since they can be more conveniently obtain in some cases. In \autoref{examples}d, our model predicts another disconnection site and adds corresponding motifs based on the predicted synthons. The predictions are also correct checked by chemists, as the prediction and ground truth differ only in the disconnection order from multi-step retrosynthesis perspective. These examples illustrate that our method can inherently learn underlying reaction rules and provide predictions with high chemical rationality.
\section{Conclusion}
Motivated by the classical retrosynthesis theory proposed by Nobel Prize laureate E.J.Corey, we have proposed a graph generative model MARS for retrosynthetic anaylsis. Our model benefits from the flexibility and low prediction complexity of motifs. The end-to-end architecture enables our model to explore latent relationships between reaction centers and motifs. Moreover, the motifs are functional groups in chemistry. It is very reasonable to treat them as elementary entities in retrosynthetic predicition task. All these account for why the high accuracy and excellent generalization performance are obtained by our model. In the future, pre-training a model to learn more reasonable motifs from existing chemical compounds will be tried.

Our work and the existing retrosynthesis methods share the limitation that lacking a more reasonable evaluation metric. Our work is capable of obtaining multiple chemically plausible reactants to synthesize products, but existing evaluation metrics only consider a given reaction. 

\section*{Broader Impact}

Our work combines end-to-end architecture and rationally designed motifs to retrosynthesis prediction, and it achieves impressive performance.  Retrosynthesis prediction is an important step in drug discovery. Our work provides a useful idea for related research, and convenience to chemists and the pharmaceutical industry. In our opinion, our method does not have a negative impact on society.

\bibliographystyle{plain}
\bibliography{mars}

\newpage



\section{Appendix A}

\subsection{Dataset Information}
\label{data_information}

The USPTO-50K dataset contains 50K reaction across 10 classes. We show that distribution of reaction classes in \autoref{tab:distribution_reaction_type}. The class imbalanced problem makes retrosynthesis prediction challenging.

\begin{table}[ht]
\centering
\caption{Distribution of 10 recognized reaction types}
\label{tab:distribution_reaction_type}
\begin{tabular}{@{}clc@{}}
\toprule
Reaction type & Reaction type name                     & \# Examples \\ \midrule
1             & Heteroatom alkylation and arylation    & 15204       \\
2             & Acylation and related processes        & 11972       \\
3             & C-C bond formation                     & 5667        \\
4             & Heterocycle formation                  & 909         \\
5             & Protections                            & 672         \\
6             & Deprotections                          & 8405        \\
7             & Reductions                             & 4642        \\
8             & Oxidations                             & 822         \\
9             & Functional group interconversion (FGI) & 1858        \\
10            & Functional group addition (FGA)        & 231         \\
\bottomrule
\end{tabular}
\end{table}

\subsection{Atom and Bond features}

\label{atom_bond_features}

We follow \cite{ydz:retroxpert} to design atom and bond features. All the features are computed using rdkit.

\begin{table}[ht]
\centering
\caption{Atom Feature used in MARS. All features are one-hot encoding, except the atomic mass is a real number scaled to be on the same order of magnitude.}
\label{tab:atom_feature}
\begin{tabular}{@{}lcc@{}}
\toprule
Feature       & Description                                      & Size \\ \midrule
Atom type     & Type of atom (i.e. C, N, O), by atomic number.   & 17   \\
\# Bond       & Number of bonds the atom is involved in.         & 7    \\
Formal charge & Interger electronic charge assigned to atom.     & 5    \\
Chirality     & Unsoecified, tetrahedral CW/CCW, or other.       & 4    \\
\# Hs         & Number of bonded Hydrogen atom.                  & 5    \\
Hybridization & sp, sp2, sp3, sp3d, or sp3d2.                    & 5    \\
Aromaticity   & Whether this atom is part of an aromatic system. & 1    \\
Atomic mass   & Mass of the atom, divided by 100.                & 1    \\
\midrule
Reaction type & The specified reaction type if it exists.        & 10   \\
\bottomrule
\end{tabular}
\end{table}

\begin{table}[ht]
\centering
\caption{Bond Feature used in MARS. All features are one-hot encoding.}
\label{tab:bond_feature}
\begin{tabular}{@{}lcc@{}}
\toprule
Feature     & \multicolumn{1}{c}{Description}      & \multicolumn{1}{c}{Size} \\ \midrule
Bond type   & Single, double, triple, or aromatic. & 4                        \\
Conjugation & Whether the bond is conjugated.      & 1                        \\
In ring     & Whether the bond is part of a ring.  & 1                        \\
Stereo      & None, any, E/Z or cis/trans.         & 6                        \\ \bottomrule
\end{tabular}
\end{table}

\subsection{Experimental Setting} 

\label{experimental_setting}

We use PyTorch \cite{paszke2019pytorch} and PyG \cite{fey2019pyg} to implement our model. We use Graph Transformer \cite{shi:graphtransformer} as our graph encoder, which stacked six eight-head attentive layers. The GRU network is implemented with three layers. Embedding size \(D\) in our model is set to \(512\). We show the parameter settings for other modules in the \autoref{tab:module_setting}. In the hidden layer of these modules, we use Mish \cite{misra2019mish} as the activation function, defined as:
\begin{equation}
    \begin{aligned}
        \operatorname{Mish}(x) = x\times \operatorname{tanh}(\log(1+e^{x}))
    \end{aligned}
\end{equation}

In all experiments, we train on USPTO-50K for 100 epochs, using a batch size of \(32\) and the Adam \cite{kingma2014adam} optimizer with initial learning rate of \(0.0003\). We use the strategy of cosine annealing learning rate with restart \cite{loshchilov2016sgdr} to accelerate convergence and avoid local optima, which the restart cycle is set to \(20\) epochs. The training on USPTO-50K takes approximately \(17\)h on a single NVIDIA Tesla V100 GPU. The beam size \(k\) is set to \(10\) in inference phase. 

\begin{table}[ht]
\centering
\caption{Parameter settings for each module in MARS}
\label{tab:module_setting}
\begin{tabular}{@{}lccccc@{}}
\toprule
Modules                                            & Input size & Output size & \# layer & Dropout rate & Activation Function \\ \midrule
\(\sigma_G(\cdot)\)                           & 512        & 1536     & 1        & 0.3          & -                   \\
\(\sigma_e(\cdot)\)                           & 512        & 512         & 1        & 0            & -                   \\
\(\sigma_b(\cdot)\)                           & 4          & 512         & 1        & 0            & -                   \\
\(\sigma_{atom}(\cdot)\)                    & 512        & 512         & 1        & 0            & -                   \\
\(\sigma_z(\cdot)\)                           & 300        & 512         & 1        & 0            & -                   \\
\(\operatorname{MLP}_{bond}(\cdot)\)      & 1024       & 512         & 1        & 0            & Mish                \\
\(\operatorname{MLP}_{act}(\cdot)\)       & 1024       & 8           & 1        & 0            & Mish                \\
\(\operatorname{MLP}_{target}(\cdot)\)    & 1536       & 1           & 2        & 0.3          & Mish                \\
\(\operatorname{MLP}_{type}(\cdot)\)      & 1536       & 4           & 2        & 0            & Mish                \\
\(\operatorname{MLP}_{motif}(\cdot)\)     & 1024       & 211         & 2        & 0.4          & Mish                \\
\(\operatorname{MLP}_{interface}(\cdot)\) & 1536       & 4           & 2        & 0            & Mish                \\ \bottomrule
\end{tabular}
\end{table}

\subsection{Baselines} 

\label{baselines}

We take three template-based and seven template-free methods as our competitors. 

For template-based models, RetroSim \cite{coley:retrosim} selects reaction centers based on Morgan fingerprint similarity between target molecules and known precedents. NeuralSym \cite{segler:neuralsym} combines a fully-connect layer and a deep highway network to learn knowledge of potential correlations between molecular functional groups and reactions. GLN \cite{dai:gln} models joint probability of single-step retrosynthesis to select templates and generate reactants. 

Template-free models can be divided into three sequence-based models and four graph-based models. For sequence-based models, SCROP \cite{zheng:scrop} combines an extra Transformer to correct predicted SMILES strings. LV-Transformer \cite{chen:lvtransformer} uses a pre-training strategy and introduces latent variables to improve prediction diversity. and DualTF \cite{sdl:dualtf} unifies sequence-based and graph-based models using energy functions and uses an extra order model to help inference. 

For graph-based models, G2Gs\cite{shi:g2gs} employs a graph neural network to select reaction centers and generates reactants using a variational autoencoder. RetroXpert \cite{ydz:retroxpert} leverages a graph neural network to predict disconnections and regards reactant generation as a sequence translation task. GraphRetro \cite{srb:graphretro} determines the synthon through an edit prediction model, and then perform a single full-connected network  to complete the synthons by using predefined leaving groups. MEGAN \cite{sbb:megan} defines five graph editing actions, using two stacked graph attention networks to perform retrosynthesis predictions.

\newpage
\section{Appendix B}

\subsection{Ablation Study of Graph Encoder}
Graph encoder can learn the representations of nodes by aggregating their neighbors' information, which is used to learn the topological information of molecular graphs. In \autoref{tab:ablation_encoder}, we investigate the effect of different graph encoders on the  performance, including GCN \cite{kipf2016semi}, GAT \cite{velivckovic2017graph}, GraphSAGE\cite{hamilton2017inductive} and GTN\cite{shi:graphtransformer}. GTN outperforms other graph encoders and is set as the default graph encoder for MARS.
\begin{table}[ht]
\centering
\caption{Top-k accuracy of different graph encoders.}
\label{tab:ablation_encoder}
\begin{tabular}{@{}lcccccccc@{}}
\toprule
\multirow{3}{*}{Graph encoder} & \multicolumn{8}{c}{Top-k Accuracy (\%)}                                                                                               \\ \cmidrule(l){2-9} 
                                     & \multicolumn{4}{c}{Reaction Type known}                           & \multicolumn{4}{c}{Reaction Type Unknown}                         \\ \cmidrule(l){2-9} 
                                     & 1              & 3              & 5              & 10             & 1              & 3              & 5              & 10             \\ \midrule
GCN                                  & 0.632          & 0.847          & 0.894          & 0.925          & 0.498          & 0.721          & 0.798          & 0.861          \\
GAT                                  & 0.646          & 0.852          & 0.897          & 0.922          & 0.516          & 0.722          & 0.794          & 0.854          \\
GraphSAGE                            & 0.644          & 0.845          & 0.892          & 0.923          & 0.522          & 0.738          & 0.808          & 0.867          \\
GTN                                  & \textbf{0.662} & \textbf{0.858} & \textbf{0.902} & \textbf{0.929} & \textbf{0.546} & \textbf{0.764} & \textbf{0.833} & \textbf{0.885} \\ \bottomrule
\end{tabular}
\end{table}

\subsection{Ablation Study of Representation Pooling Function}
The representation pooling function is utilized to aggregate all atom representations to obtain the molecular graph representation \(h_G\). We compare the effects of several pooling functions on performance, including max, sum, mean and attention pooling. As shown in \autoref{tab:ablation_pooling}, attention pooling achieves the best performance for learning the representation of molecular graphs.
\begin{table}[ht]
\centering
\caption{Top-k accuracy of different pooling functions.}
\label{tab:ablation_pooling}
\begin{tabular}{@{}lcccccccc@{}}
\toprule
\multirow{3}{*}{\makecell[l]{Representation\\ pooling function}} & \multicolumn{8}{c}{Top-k Accuracy (\%)}                                                                                               \\ \cmidrule(l){2-9} 
                                     & \multicolumn{4}{c}{Reaction Type known}                           & \multicolumn{4}{c}{Reaction Type Unknown}                         \\ \cmidrule(l){2-9} 
                                     & 1              & 3              & 5              & 10             & 1              & 3              & 5              & 10             \\ \midrule
Max                                  & 0.576          & 0.807          & 0.865          & 0.903          & 0.381          & 0.603          & 0.696          & 0.765          \\
Sum                                  & 0.568          & 0.795          & 0.850           & 0.891          & 0.436          & 0.673          & 0.751          & 0.815          \\
Mean                                 & 0.643          & 0.849          & 0.897          & 0.926          & 0.530           & 0.755          & 0.822          & 0.880           \\
Attention                            & \textbf{0.662} & \textbf{0.858} & \textbf{0.902} & \textbf{0.929} & \textbf{0.546} & \textbf{0.764} & \textbf{0.833} & \textbf{0.885} \\ \bottomrule
\end{tabular}
\end{table}

\subsection{Ablation Study of Embedding Size}
We compare the effect of several embedding size \(K\) on the performance of MARS. The result is shown in \autoref{tab:ablation_dimension}.  When the embedding size is 512, the top-\(1\) accuracy is optimal. However, when embedding size is set to 1024, the performance is slightly degraded. Therefore, we choose \(512\) as our default embedding size.
\begin{table}[!ht]
\centering
\caption{Top-k accuracy of different embedding sizes.}
\label{tab:ablation_dimension}
\begin{tabular}{@{}lcccccccc@{}}
\toprule
\multirow{3}{*}{Embedding size} & \multicolumn{8}{c}{Top-k Accuracy (\%)}                                                                                               \\ \cmidrule(l){2-9} 
                                     & \multicolumn{4}{c}{Reaction Type known}                           & \multicolumn{4}{c}{Reaction Type Unknown}                         \\ \cmidrule(l){2-9} 
                                     & 1              & 3              & 5              & 10             & 1              & 3              & 5              & 10             \\ \midrule
64                                   & 0.597          & 0.835          & 0.889          & 0.923          & 0.480          & 0.730          & 0.812          & 0.872          \\
128                                  & 0.628          & 0.856          & \textbf{0.903} & \textbf{0.934} & 0.514          & 0.757          & 0.828          & 0.887          \\
256                                  & 0.642          & 0.854          & \textbf{0.903} & 0.931          & 0.526          & 0.759          & 0.831          & \textbf{0.888} \\
512                                  & \textbf{0.662} & \textbf{0.858} & 0.902          & 0.929          & \textbf{0.546} & \textbf{0.764} & \textbf{0.833} & 0.885          \\
1024                                 & 0.637          & 0.848          & 0.893          & 0.926          & 0.526          & 0.739          & 0.807          & 0.860          \\ \bottomrule
\end{tabular}
\end{table}

\end{document}